\documentclass[10pt,twocolumn,letterpaper]{article}

\usepackage[pagenumbers]{cvpr} 

\definecolor{cvprblue}{rgb}{0.21,0.49,0.74}
\usepackage[pagebackref,breaklinks,colorlinks,allcolors=cvprblue]{hyperref}

\usepackage{transparent}
\usepackage{color}
\usepackage{xcolor}
\usepackage{wrapfig}
\usepackage{colortbl}
\usepackage[makeroom]{cancel}
\usepackage{soul}  
\usepackage{pifont}
\usepackage{rotating} 

\definecolor{white}{rgb}{1,1,1}
\definecolor{gray}{rgb}{0.5,0.5,0.5}
\definecolor{darkergreen}{RGB}{21, 152, 56}
\definecolor{darkerred}{RGB}{220, 35, 120}
\definecolor{darkerblue}{rgb}{0,0.08,0.45} 
\definecolor{royalblue}{RGB}{65,105,225}
\definecolor{lightblue}{RGB}{231,247,255}
\definecolor{gray94}{gray}{.94}
\definecolor{gray90}{gray}{.90}

\newcommand{\gray}[1]{\textcolor{gray}{#1}}

\definecolor{plblue}{RGB}{100,59,237}

\newcommand{\grow}[1]{\cellcolor{gray94}{#1}}

\definecolor{colorA}{RGB}{229,166,175}  
\colorlet{colorA}{colorA!210}
\definecolor{colorB}{RGB}{154,207,241}
\colorlet{colorB}{colorB!260}
\definecolor{colorC}{RGB}{158,218,200}
\colorlet{colorC}{colorC!280}
\definecolor{colorD}{RGB}{195,184,234}
\colorlet{colorD}{colorD!250}
\newcommand{\xmark}{\ding{55}}%
\newcommand{\xmarkg}{\textcolor{gray}{\ding{55}}}%

\def\paperID{10741} 
\def\confName{CVPR}
\def\confYear{2025}

\title{MergeVQ: A Unified Framework for Visual Generation and Representation with Disentangled Token Merging and Quantization}

\author{
\bf{Siyuan Li}$^{1,3*}$~~~~Luyuan Zhang$^{2*}$~~~~Zedong Wang$^{4}$~~~~Juanxi Tian$^{3}$~~~~Cheng Tan$^{1,3}$\\
\bf{Zicheng Liu}$^{1,3}$~~~~Chang Yu$^{3}$~~~~Qingsong Xie$^{5\dagger}$~~~~Haonan Lu$^{5}$~~~~Haoqian Wang$^{2}$~~~~Zhen Lei$^{6,7,8\dagger}$\\
$^1$Zhejiang University~~~~~$^2$Tsinghua University~~~~~$^3$Westlake University\\
$^4$The Hong Kong University of Science and Technology~~~~~$^5$OPPO AI Center\\
$^6$CAIR, HKISI-CAS~~~~~$^7$MAIS CASIA~~~~~$^8$University of Chinese Academy of Sciences\\
}

\begin{document}
\maketitle

\begin{abstract}

Masked Image Modeling (MIM) with Vector Quantization (VQ) has achieved great success in both self-supervised pre-training and image generation. However, most existing methods struggle to address the trade-off in shared latent space for generation quality \textit{vs.} representation learning and efficiency. To push the limits of this paradigm, we propose MergeVQ, which incorporates token merging techniques into VQ-based autoregressive generative models to bridge the gap between visual generation and representation learning in a unified architecture. During pre-training, MergeVQ decouples top-k semantics from latent space with a token merge module after self-attention blocks in the encoder for subsequent Look-up Free Quantization (LFQ) and global alignment and recovers their fine-grained details through cross-attention in the decoder for reconstruction. As for the second-stage generation, we introduce MergeAR, which performs KV Cache compression for efficient raster-order prediction. Experiments on ImageNet verify that MergeVQ as an AR generative model achieves competitive performance in both representation learning and image generation tasks while maintaining favorable token efficiency and inference speed.
The source code will be available at \url{https://apexgen-x.github.io/MergeVQ}.


\end{abstract}


\begin{NoHyper}
\def\thefootnote{*}\footnotetext{Equal contribution.~~~~~~ $^\dagger$Corresonding authors.}
\end{NoHyper}
\def\thefootnote{\arabic{footnote}}
\section{Introduction}
\label{sec:intro}

Vector Quantization (VQ)~\cite{2017VQ-VAE} has garnered increasing attention for its ability to encode continuous visual signals into discrete tokens, enabling autoregressive (AR) models to process visual modalities. Since VQGAN~\cite{cvpr2021vqgan},  most visual AR generative models have adopted a two-stage design: first encode signals into discrete latent space for pre-training, then generate them with an autoregressive Transformer. Besides generation, BEiT~\cite{iclr2022BEiT} proposed Masked Image Modeling (MIM) based on the VQ framework, achieving successful latent-based pretraining~\cite{eccv2022mcBEiT, cvpr2023MAGE} and thus attracting growing interest in unifying visual representation learning and generation tasks in a \textit{shared latent space}~\cite{zhou2021ibot}.

\begin{figure}[t!]
    \vspace{-0.75em}
    \centering
    \includegraphics[width=1.0\linewidth]{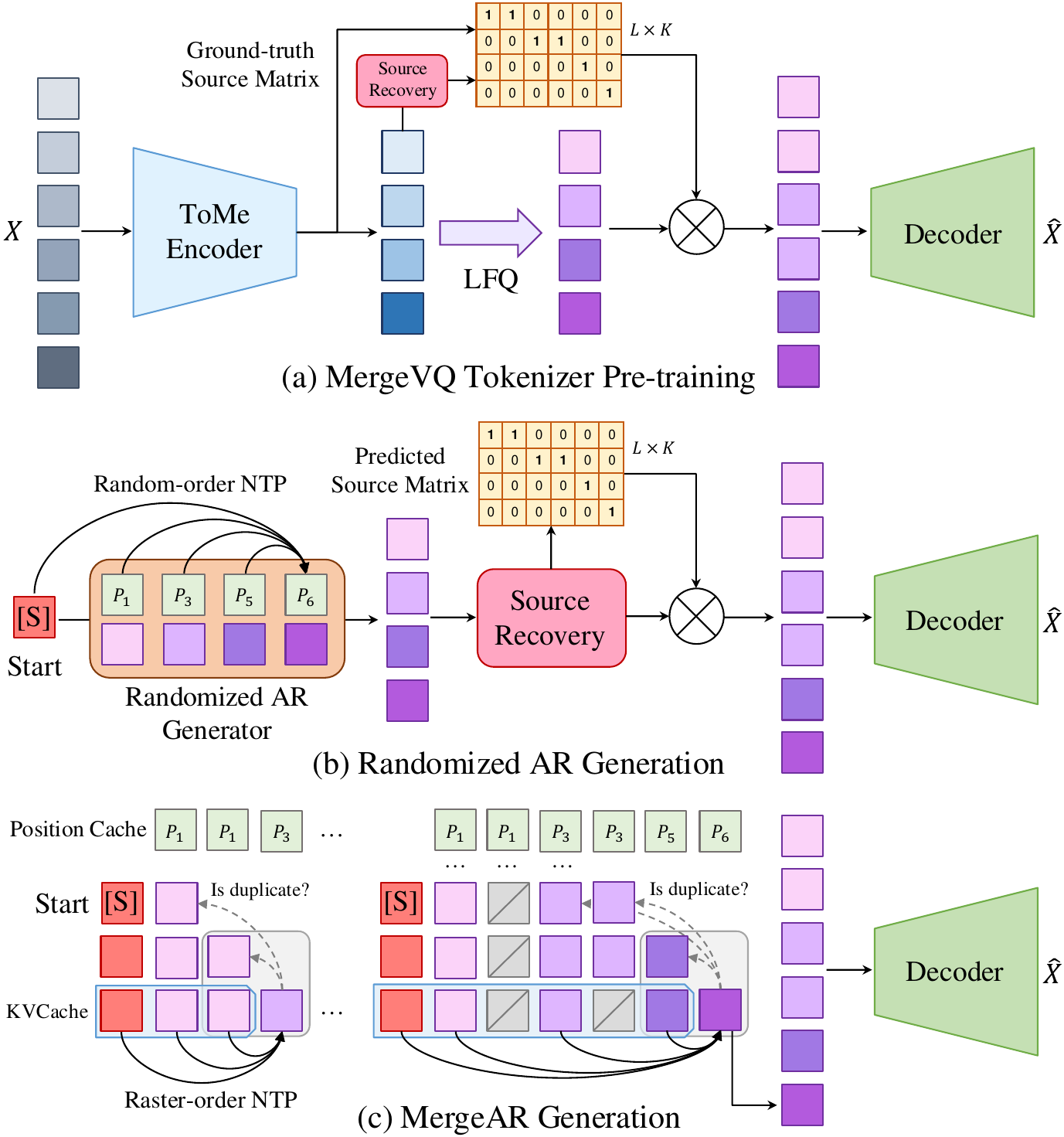}
    \vspace{-1.5em}
    \caption{\textbf{MergeVQ learning paradigms}. \textbf{(a)} The MergeVQ Tokenizer extracts $K$ semantic tokens with decoupled positional information (retained in the source matrix) by ToMe~\cite{iclr2022ToMe} while quantizing spatial details by LFQ~\cite{iclr2024FSQ, ICLR2024magvit2}, which will be recovered and reconstructed correspondingly. \textbf{(b)} MergeVQ with random-order Generator~\cite{cvpr2025RandAR} generates $K$ discrete tokens with associated position instructions while trained Source Prediction and decoder restore position details. 
    \textbf{(c)} MergeAR Generator predicts $L$ tokens efficiently in a raster-order with tailored KV Cache compression to remove the redundancy within Next-token Prediction (NTP)~\cite{NIPS2024LLaMAGen}.
    }
    \label{fig:mergevq_intro}
    \vspace{-1.5em}
\end{figure}

However, recent studies~\cite{cvpr2024V2Ltokenizer, cvpr2024UnifiedIO2} have shown that visual generation and representation capabilities often lack consistency~\cite{ICLR2023MagViT2} under a VQ-based learning framework, \ie, improvements in one task may not necessarily benefit the others. This inconsistency is conjectured to arise from the competing objectives for identical embedding space: \textbf{representation learning tasks emphasize inter-class discrimination to maximize high-level semantics, while generative tasks prioritize the reconstruction of details}. In addition, training obstacles brought by VQ itself further limit the optimization process. For example, the gradient approximation in canonical VQ (\eg, VQGAN) sets an optimization bottleneck for the first-stage training. Moreover, the quantization of embedding space inevitably strips away fine-grained spatial information, which requires the models to reconstruct images with the loss of details and thus affects both the representation learning and generation.

As such, efforts have been made to extract rich semantic features from visual signals for quantization to improve the representation capacity of generative models~\cite{nips2024digit2, NIPS2024VQKD}. However, these coarse-grained semantics often sacrifice detailed information, making it difficult to support high-quality image reconstruction and generation, resulting in significant performance degradation. In this paper, we argue that representation learning and generation are not completely conflicting but with intrinsic complementarity. The crux lies in exploiting such complementarity while minimizing the information loss, which requires specific designs. To achieve this, we propose to \textit{decouple coarse-grained semantics from latent space during training and recover them for reconstruction to meet the different needs while minimizing the information loss and overhead}. By leveraging token merging techniques~\cite{iclr2022ToMe}, the encoder compresses latent space into $K$ semantic tokens while preserving the fine-grained spatial information as positions within a source matrix, as illustrated in Figure~\ref{fig:mergevq_intro}. During reconstruction, the latent fine-grained details can be restored with this source matrix, while the $K$ compressed tokens serve as high-level semantics for global alignment~\cite{iccv2021dino, zhou2021ibot}. Based on this intuition, we propose MergeVQ, which employs token merging and Look-up Free Quantization (LFQ) for spatial and channel compression.
Extensive experiments show that MergeVQ as an AR generative model achieves competitive performance in both image generation and representation learning with favorable efficiency. Our contributions can be summarized as:

\begin{itemize}[leftmargin=1.25em]
\vspace{-0.5em}
    \item We present a fresh learning paradigm that integrates token merging into a VQ-based AR generation framework, where high-level semantics are decoupled from patients in the first-stage training and can be restored with source matrix for details reconstruction, thus effectively reducing information loss while bridging the gap between representation learning and generation in a unified model.
    \item We offer two schemes for MergeVQ's second-stage generation. (i) We propose MergeAR, which performs KVCache compression for efficient raster-order prediction. (ii) With the source recovery module, existing random-order generators can also be directly used for generation.
    \item Experiments show MergeVQ's competitive performance in both visual representation learning and image generation, with favorable token efficiency and inference speed.
\end{itemize}

\section{Related Work}
\label{sec:related_work}

\subsection{Auto-regressive Image Generation}
 \label{sec:background_AR}
\textbf{Vector Quantization Tokenizer.}\quad Vector quantization, pioneered by VQ-VAE~\cite{2017VQ-VAE} and enhanced by VQ-GAN~\cite{cvpr2021vqgan} through adversarial training and Transformer integration, faces three key challenges in traditional \textit{cluster-based} approaches:
(\romannumeral1) Gradient approximation: The straight-through estimator introduces imprecise encoder gradients, an issue mitigated through extended training in MAGVIT-v2~\cite{ICLR2023MagViT2} and OpenMAGVIT2~\cite{luo2024Open-Magvit2}. 
(\romannumeral2) Inefficient codebook usage: The commitment loss often leads to uneven gradient distributions and codebook collapse. Solutions include priors regularization in RegVQ~\cite{Zhang2023RegularizedVQ} and Kepler Codebook~\cite{ICML2024keplercodebook}, and EMA normalization in BEiT.v2~\cite{2022BEiTV2} and ViT-VQGAN~\cite{ICLR2021VIT-VQGAN}.
(\romannumeral3) Discrete representation bottleneck. VQ discards fine-grained details, hindering reconstruction fidelity. RQ~\cite{CVPR2022RQVAE} addresses this through hierarchical quantization to preserve information.
\textit{Look-up Free Quantization} performs channel-wise quantization, improving codebook usage while reducing overhead. Attempts such as FSQ~\cite{iclr2024FSQ}, MAGVIT-v2~\cite{ICLR2023MagViT2}, OpenMAGVIT2~\cite{luo2024Open-Magvit2}, and advanced variants~\cite{Zhao2024ImageAV, khalil2023LL-VQ-VAE, weber2024Maskbit} demonstrate results that exceed vanilla VQ. Another line reduces inference latency with \textit{Adaptive-Length Quantization} to reduce the number of vision tokens by queries with cross-attention~\cite{nips2024titok, duggal2024Alit}, attention-based token pruning~\cite{CVPR2023MQVAE}, or token grouping strategies~\cite{duan2023QARV}.

\textbf{Autoregressive Generation.}\quad
VQGAN introduced AR visual generation by adopting the raster-order Next Token Prediction (NTP) in GPT~\cite{Radford2018GPT1, Radford2019GPT2}. Subsequent works, including LlamaGen~\cite{NIPS2024LLaMAGen} and OpenMAGVIT2~\cite{luo2024Open-Magvit2}, have extended this paradigm. In parallel, studies have focused on accelerating generation through non-autoregressive decoding, \eg, MaskGiT variants~\cite{CVPR2022maskgit, iclr2025HaltonMaskGIT} and MAR~\cite{NIPS2024MAR}, which leverages masked prediction for parallel inference. Recent advancements explore random-order AR generation, where positions are predicted prior to token embeddings (RandAR~\cite{cvpr2025RandAR}) or learnable positional encodings are utilized for prediction (RAR~\cite{Yu2024RAR}).

\begin{figure*}[t!]
    \vspace{-0.5em}
    \centering
    \includegraphics[width=1.0\textwidth]{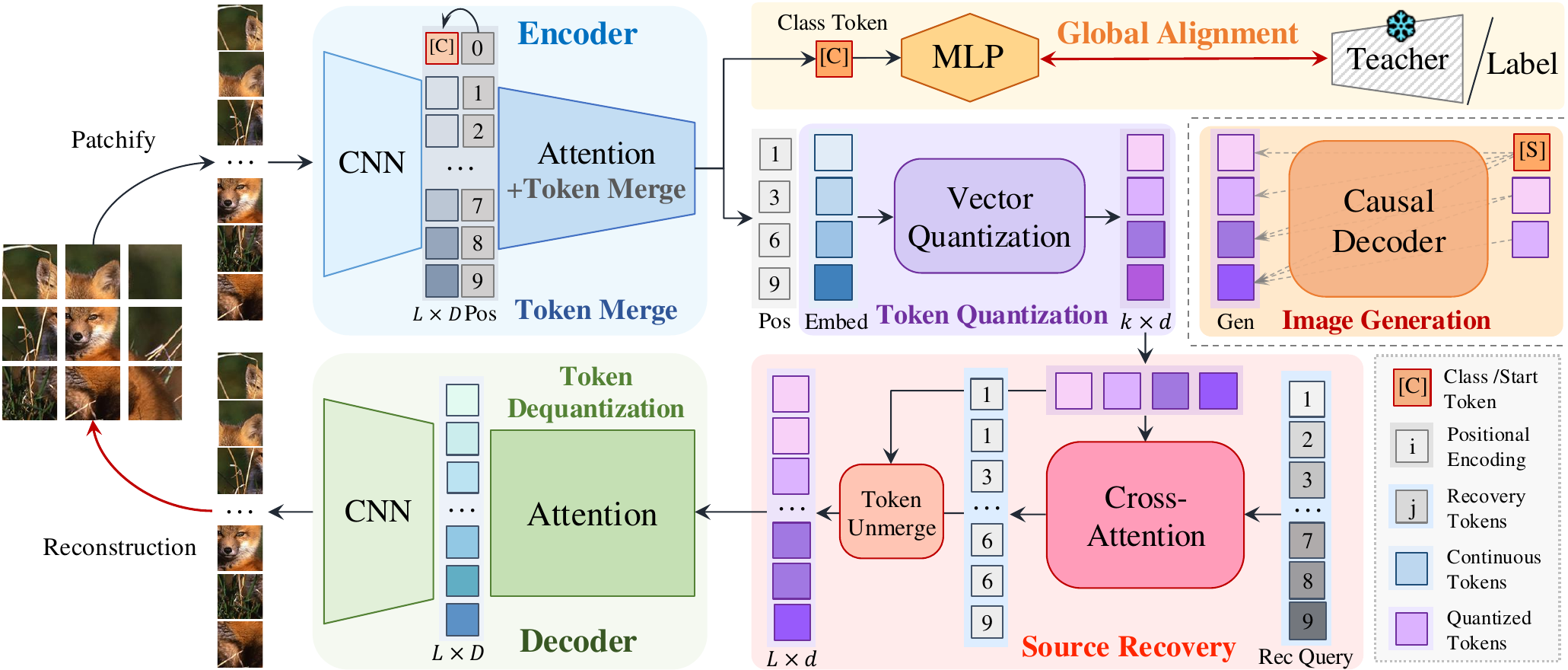}
    \vspace{-1.5em}
    \caption{\textbf{Overview of MergeVQ framework}, which contains two stages and three groups of subtasks (Sec.~\ref{sec:mergevq_framework}). \textbf{(a)} As for representation learning (Sec.~\ref{sec:mergevq_rec+rep}), $K$ semantic tokens are extracted by the encoder with self-attention and token merging~\cite{iclr2022ToMe}, which can be aligned globally with a pre-trained teacher while learning contextual information by predicting the source matrix.
    \textbf{(b)} As for reconstruction (Sec.~\ref{sec:mergevq_rec}), taking $K$ merged and quantized tokens as the input, the positional information can be retained by the Source Recovery module, and then high-quality details will be reconstructed. \textbf{(c)} As for generation (Sec.~\ref{sec:generation}), we utilize the source matrix to construct a causal mask for training and leverage the KV cache to prune repeated tokens during inference for efficient generation.
    }
    \label{fig:framework}
    \vspace{-0.5em}
\end{figure*}

\subsection{Unifying Representation and Generation}
\label{sec:background_rep+gen}

Since BEiT~\cite{iclr2022BEiT} first combined Masked Modeling with VQ for pre-training, research unifying representation and generation within a latent space has gained increasing interest~\cite{li2023MIMSurvey}. These studies, typically conducted within \textit{cluster-based VQ} frameworks, fall into two categories:
\textit{(\romannumeral1) Using Pre-training Techniques in Quantized Space.}
MQ-VAE~\cite{CVPR2023MQVAE} quantizes semantic tokens by masking important ones for reconstruction. MAGE performs Masked Modeling directly in latent space during second-stage generation training, while BEiT abandons second-stage generation, using Masked Modeling as the second stage itself.
\textit{(\romannumeral2) Using representative tasks to enhance generation quality.} DiGIT~\cite{nips2024digit2} extracts semantic tokens from pre-trained models for representation learning while using a finely crafted decoder for generation. VQ-KD~\cite{NIPS2024VQKD} employs a pre-trained teacher model to guide token reconstruction.
REPA~\cite{Yu2024REPA} proposes that representation alignment can significantly improve the training efficiency and generation quality of diffusion models.
Some approaches align visual and text codebooks via CLIP-inspired methods~\cite{zhang2024VAR-CLIP}. SPAE~\cite{nips2023SPAE} utilizes hierarchical codebooks to align visual representations with frozen LLMs, while V2L Tokenizer~\cite{cvpr2024V2Ltokenizer} employs both global and local tokenizers for multi-modal alignment.

\subsection{Token Compression in Transformer}
\label{sec:background_compression}

Token compression techniques have emerged as crucial components for improving efficiency in Transformer-based architectures, particularly in ViTs and LLMs. As for the Transformer encoder, ToMe variants \cite{acl2023pumer, cvpr2023tm4sd, arxiv2023ltmp, chen2023diffrate} employ lightweight bipartite soft matching (BSM) to achieve pruning-like efficiency gains, enhancing ViT throughput with minimal performance degradation. However, BSM-based methods often incur information loss among tokens due to their heuristic merging rules. Clustering-based token merging strategies, including k-means~\cite{Marin_2023_k_means} and spectral clustering~\cite{2020pmlrSpectral_Clustering}, have been explored to address this issue through more controllable operations. Yet, these techniques introduce computationally intensive iterative protocols in ViT layers.
As for decoder architectures, recent advancements in KV cache compression (\eg, StreamLLM~\cite{iclr2024StreamLLM}, FastGen~\cite{nips2024fastgen}, SnapKV~\cite{nips2024snapKV}, and H\(_2\)O~\cite{nips2023h2o}) propose to optimize memory usage and inference speed via selective token retention and key-value pair compression. While these methods significantly enhance LLM inference efficiency, they are not directly applicable in the training phase.
%

\section{MergeVQ Learning Paradigm}
\label{sec:method}

\subsection{MergeVQ Framework}
\label{sec:mergevq_framework}
This section introduces MergeVQ, a VQ-based visual representation learning and auto-regressive image generation framework, and formalizes its core components.

\textbf{Token Merge Encoding}: Given an input image $X \in \mathbb{R}^{H \times W \times 3}$, we employ a two-stage encoder $\mathcal{E}_{\phi,\theta} (\cdot)$ for feature extraction. First, a CNN encoder $\mathcal{E}_{\phi} (\cdot)$ extracts feature map $Z \in \mathbb{R}^{\frac{H}{f} \times \frac{W}{f} \times D}$, where $f$ is the downsampling factor and $d$ denotes the channel dimension. This feature is then flattened into a $L$-length token sequence $Z_L \in \mathbb{R}^{L \times D}$ as: 
\begin{equation}
    Z_L = \mathcal{E}_{\phi}(X).
    \label{cnn}
\end{equation}
In the second stage, we employ an attention-based encoder with token merging modules~\citep{iclr2022ToMe}, denoted as $\mathcal{E}_{\theta} (\cdot)$, to further compress $Z_L$ into condensed $K$-length tokens $Z_K \in \mathbb{R}^{K \times D}$ alongside a source matrix $S \in \mathbb{R}^{K \times L}$ that encodes spatial relationships between merged and original tokens:


\begin{equation}
    S, Z_K= \mathcal{E}_{\theta}(Z_L).
    \label{equ:tome_encoder}
\end{equation}
The whole encoding process of MergeVQ is thus as:

\begin{equation}
    S, Z_K = \mathcal{E}_{\phi,\theta}(X).
\end{equation}
To ensure that $Z_K$ retains rich high-level semantics, we also impose global alignment constraints discussed in Sec.~\ref{sec:mergevq_rec+rep}.

\textbf{Quantization}: We adopt \textit{LFQ} to discretize the merged latent $Z_{K}$. Concretely, the codebook $\mathcal{C}$ comprises binary vectors defined as:
$
    \mathcal{C} = \times_{i=1}^d \{-1, 1\}, \quad |C| = 2^d,
$
where $d$ is the quantized dimension. As such, each token $z_{Ki} \in Z_K$ is quantized element-wise:
$z_{Ki} = \text{sign}(z_{Ki}) = -1\cdot \mathbb{I}(z_{Ki} < 0)+ \mathbb{I}(z_{Ki} > 0)$.
Then, the index of quantized feature $z_{mi}$ is computed as a binary integer:
$\text{Index}(z_{Ki}) = \sum_{j=1}^N 2^{k-1} \cdot \mathbb{I}(z_{Kij} > 0)$, yielding quantized tokens $\tilde{Z}_{K}$ as:
\begin{equation}
    \tilde{Z}_{K} = \mathcal{Q}(Z_K,\mathcal{C}),
\end{equation}

\textbf{Token Recovery and Reconstruction}: The key design lies in exploiting the spatial priors in source matrix $S$, which inherently encodes fine-grained positional dependencies between original $L$-length tokens and compressed ones during merging. We thus propose the recovery module $ \mathcal{R}_{\omega} (\cdot, \cdot) $ to map quantized $\tilde{Z}_{K}$ back to $\tilde{Z}_{L}$ with the original length:
\begin{equation}
    \tilde{Z}_{L} = \mathcal{R}_{\omega}(\tilde{Z}_{K},S).
    \label{equ:rec}
\end{equation}
This enables MergeVQ to retain both the coarse-grained semantics and fine-grained details, effectively balancing compression and reconstruction.
The recovered $\tilde{Z}_{L}$ is then decoded into pixel space by $ \mathcal{D}_{\psi} (\cdot)$ for reconstruction:
\begin{align}
   \hat{X} = \mathcal{D}_{\psi}( \tilde{Z}_{L}).
   \label{equ:decode}
\end{align}
By unifying the efficiency of ToMe with the spatial priors in $S$, MergeVQ aims to achieve \textit{loss-aware} encoding: merged tokens are not merely reduced computational overhead but retained positional information for recoverable details.

\subsection{Harmonize Representation and Generation}
\label{sec:mergevq_rec+rep}
As aforementioned, we suppose that the overlooked explicit modeling of latent token-level context might serve as a critical gap for autoregressive generation, where next-token prediction relies on coherent spatial and semantic relationships that existing VQ techniques fail to capture. To address this, we introduce an additional Source Recovery task to the first-stage learning, which trains the model to recover the token context encoded in source matrix $S$ (illustrated in Figure~\ref{fig:framework}).

\textbf{Attention with Token Merging}: Building on ToMe~\cite{iclr2022ToMe}, we iteratively merge tokens across $N$ attention layers while maintaining a binary source $S \in \{ 0, 1\}^{K \times L} $ that records the ancestry of each merged token. Given the initial sequence $Z^{(0)}_L = \mathcal{E}_{\phi} (X)$, the $l$-th ToMeAttention merges tokens as:
\begin{equation}
    S^{(l)}, Z_{\textrm{K}}^{(l)} = \textrm{ToMeAttention}^{(l)}\big( Z_{L}^{(l)},S^{(l-1)}, r\big),
\end{equation}
where $S^{(0)}=I_{L}$ and $l \in [1, N]$. Note that $Z_{L}^{(l+1)} = Z_{K}^{(l)}$ with $l\le N-1$. At each layer, the top $2r$ tokens by similarity score are merged into $r$ tokens, reducing sequence length to $K=L-rN$ after $N$ layers as Eq.~(\ref{equ:tome_encoder}). As such, $S$ inherently preserves the positional information of merged tokens $Z_K$ during encoding, which enables subsequent recovery. Please view Appendix~\ref{app:network} for implementation details.

\begin{figure}[t!]
    \centering
    \includegraphics[width=1.0\linewidth]{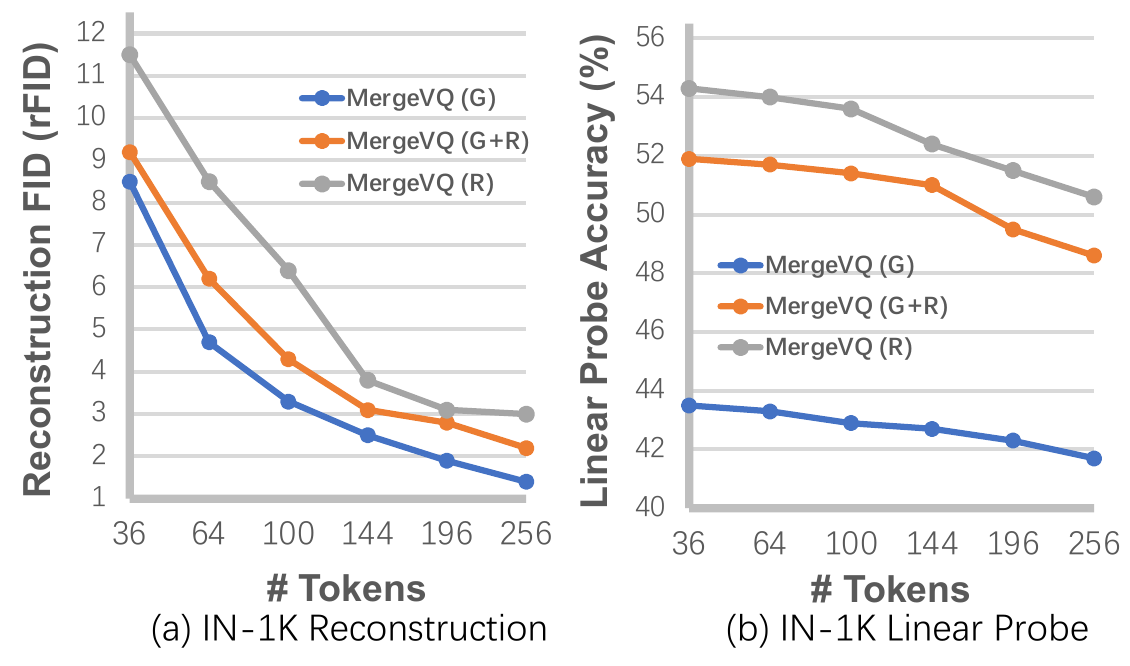}
    \vspace{-2.0em}
    \caption{\textbf{Analysis of kept tokens in reconstruction and representation learning}. Three MergeVQ tokenizers are trained with $128$ resolution for 30 epochs on ImageNet-1K. 
    They keep 256, 144, and 36 tokens with ToMe~\cite{iclr2022ToMe} in the encoder during training. In inference, we evaluate rFID and linear probing top-1 accuracy with diverse merge ratios to show the trade-off between generation and representation. Please view Sec.~\ref{sec:exp} and Appendix~\ref{app:result} for details.
    }
    \label{fig:merge_ratio_analysis}
    \vspace{-0.75em}
\end{figure}
\begin{figure}[t!]
    \centering
    \includegraphics[width=1.0\linewidth]{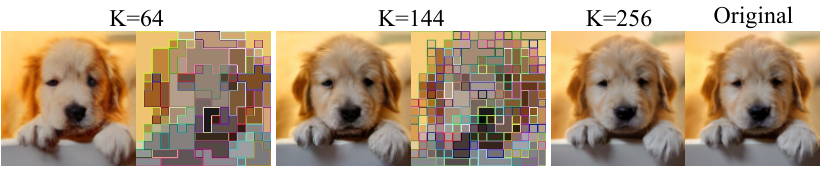}
    \vspace{-1.75em}
    \caption{\textbf{Visualization of MergeVQ (G+R) reconstruction}. With the kept tokens varying from 64 to 256, clustering maps of ToMe Attention indicate that MergeVQ can extract discriminative semantic tokens while recovering contextual positions and details.
    }
    \label{fig:vis_rec_tome}
    \vspace{-1.25em}
\end{figure}

\textbf{Source Recovery Model}:
As mentioned above, canonical VQ methods discard the contextual interactions in latent space. MergeVQ addresses this via a lightweight transformer decoder as the Source Recovery Model that learns to recover $S$ from the quantized tokens $\tilde{Z}_{K}$ since $S$ is unavailable during generation. In particular, the decoder with $L$ learnable recovery queries $Q_{r} \in \mathbb{R}^{L \times d}$ attends to $\tilde{Z}_{K}$ through cross-attention for semantics interaction. Subsequently, two self-attention layers further refine $Q_{r}$ into $\tilde{Q} \in \mathbb{R}^{L\times d}$, capturing latent token relationships.
Since the source matrix records the positional relationships between $K$-centered $\tilde{Z}_{K}$ and original ${Z}_{L}$, we use $\tilde{Z}_{K}$ as clustering centers to classify $\tilde{Q}$, which can be formulated as:
\begin{equation}
    \hat{S}^\top = \arg \max \left( \text{softmax}\left( \tilde{Q} \tilde{Z}_{K}^\top \right) \right).
\end{equation}
We employ cross-entropy as our learning objective $\mathcal{L}_{\textrm{src}}$ to measure the difference between $\hat{S}$ and $S$, as: 
\begin{equation}
    \mathcal{L}_{\textrm{src}} = - \sum_{i,j} S_{i,j} \log (\hat{S}_{i,j}) + (1 - S_{i,j}) \log (1 - \hat{S}_{i,j}).
    \label{eq:loss_src}
\end{equation}
As such, this enforces the model to \textit{internalize how tokens were merged}—a form of token-level context absent in existing VQ. During second-stage AR generation, when $S$ is inaccessible, the trained decoder infers context directly from $\tilde{Z}_{K}$, enabling accurate recovery for high-quality generation.

\textbf{Global Alignment}: To further enhance token representations for discriminative tasks, we align the merged tokens $Z_K$ with global image semantics through the self-distillation proposed by DINO~\cite{iccv2021dino}. 
We uniformly sample an image \( X \) from the training set, apply random augmentations to generate views \( u \) and \( v \), and feed them into the DINOv2 encoder $\mathcal{E}_{\theta '} (\cdot)$~\cite{Oquab2023DINOv2} and MergeVQ. The predicted category distributions from the [CLS] tokens, \( v_t = P_{\theta '}^{[CLS]}(v) \) and \( u_t = P_{\theta}^{[CLS]}(u) \), are aligned by minimizing the cross-entropy between them, which can be formulated as:
\begin{equation}
    \mathcal{L}_{[CLS]} = -P_{\theta '}^{[CLS]}(v)^\top \log P_{\theta}^{[CLS]}(u).
    \label{eq:alignment}
\end{equation}
This ensures $Z_K$ encodes semantically rich visual concepts while retaining compatibility with subsequent recovery.


\subsection{Token Recovery and Reconstruction}
\label{sec:mergevq_rec}

This section details how MergeVQ bridges token compression with high-fidelity reconstruction in first-stage training.

\textbf{Token Recovery for Reconstruction}:
As stated in Sec.~\ref{sec:mergevq_framework}, we perform token recovery to restore fine-grained positional information before reconstruction. This is achieved through the source matrix $S$ as denoted in Eq.~(\ref{equ:rec})
Specifically,
we utilize the positional information in $S$ to expand $Z_K$ back into a sequence of length $L$. For example, if the $i$-th row of $S$ satisfies $S(i, j_1) = 1$ and $S(i, j_2) = 1$, we recover the $L$-length $\tilde{Z}_L$ such that $\tilde{Z}_{Lj_1} = \tilde{Z}_{Lj_2} = \tilde{Z}_{Ki}$, which can thus be implemented as:
\begin{equation}
\begin{aligned}
    \hspace{-0.5em}
     \tilde{Z}_{L} =  [\tilde{z}_l]_{l=1}^{L} = S^\top \tilde{Z}_{K} 
    = \left[ \sum_{i=1}^K \tilde{z}_{Ki} \times s_{il} \right]_{i=1}^{L}.
    \end{aligned}
\end{equation}
Subsequently, we apply the decoder $\mathcal{D}_{\psi}$ to reconstruct the recovered $\tilde{Z}_L$ as Eq.~(\ref{equ:decode}). Note that we obtain the ground-truth source matrix during first-stage encoding, allowing straightforward token recovery. In the second phase, the predicted source matrix $\hat{S}$ could also be obtained from the pre-trained Source Recovery Model discussed in Sec.~\ref{sec:background_rep+gen}, which enables \textit{context-aware} image token expansion.

\textbf{Hybrid Model with Weight Initialization}:
Mainstream generative models typically rely on CNNs for feature extraction, while pure Transformer-based architectures are comparatively rare. However, in visual representation learning, Transformers are prevalent. MergeVQ combines these paradigms into a hybrid one: the CNN encoder $\mathcal{E}_{\phi} (\cdot)$ extracts low-level features, providing inductive bias for pixel-aligned reconstruction. Subsequent layers $\mathcal{E}_{\theta} (\cdot)$ employ Transformer with ToMeAttention for dynamic downsampling, balancing attention efficiency with representation capabilities. To further exploit these benefits, we integrate a pre-initialized Transformer into our architecture. The network details are illustrated in Figure~\ref{fig:framework} and Appendix~\ref{app:network}.

\begin{figure}[t!]
    \vspace{-0.25em}
    \centering
    \includegraphics[width=1.0\linewidth]{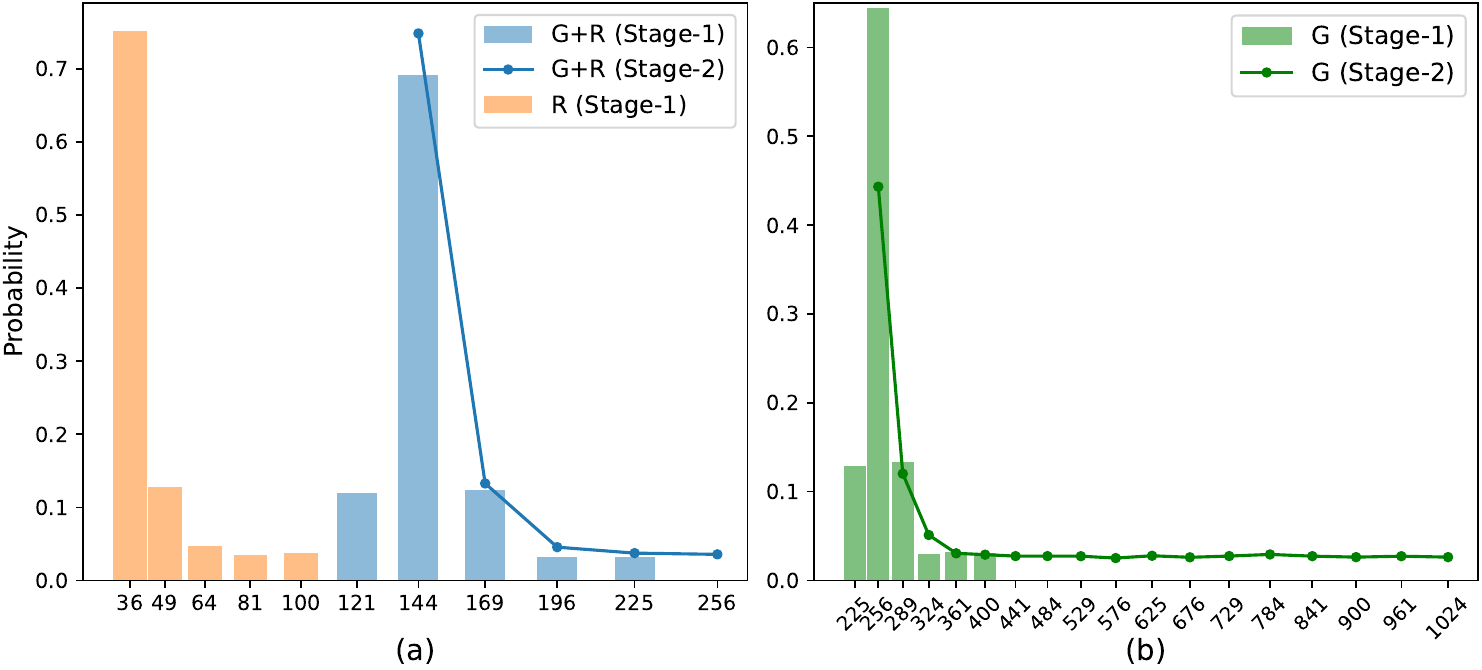}
    \vspace{-1.75em}
    \caption{\textbf{Distribution of merge ratios sampling in training}. (a) With 256 tokens in total, MergeVQ (R) and (G+R) sample the square number as kept token numbers in $[36, 100]$ and $[121, 225]$ with exponential and Gaussian distributions for stage-1 training, while the G+R version sampling from $[144, 256]$ for stage-2 training.
    (b) With 1024 tokens in total, MergeVQ (G) samples the square kept number in $[225,400]$ and $[256,1024]$ with Gaussian and exponential distributions in both stage-1 and stage-2 training.
    }
    \label{fig:merge_ratio_sampling}
    \vspace{-1.5em}
\end{figure}

\textbf{Adaptive Merge Ratios for Diverse Tasks}: 
%
Unlike existing adaptive-length quantization strategies~\cite{nips2024titok, li2024Imagefolder}, our MergeVQ utilizes variable merge ratios $r$ during training instead of fixed sequence lengths. The ToMe module provides flexibility for different tasks through adjustable merge ratios. Experiments show that representation learning and reconstruction tasks benefit from diverse merge ratio settings. For instance, as shown in Figure~\ref{fig:merge_ratio_analysis}, representation learning (Sec.~\ref{sec:mergevq_rec+rep}) favors larger merge ratios~\cite{cvpr2022MAE, iclr2023LayerGraft}, which might help capture the discriminative global patterns.
Therefore, we present three variants: the \underline{R}epresentation (R) version for enhanced generalization, the \underline{G}eneration and \underline{R}epresentation (R+G) version bridging both objectives, and the \underline{G}eneration (G) one preserving spatial fidelity for high-quality synthesis.
More importantly, we propose a merge ratio sampling strategy in Figure~\ref{fig:merge_ratio_sampling} to expose the model to varying token counts, thus further enhancing the robustness and generalization capability of MergeVQ through the two-stage training. In practice, we retained three versions of merged token counts: 256 for (G), 144 for (R+G), and 36 for (R), respectively. During training, we determine the corresponding ratio $r$ by sampling the number of tokens retained, focusing on a range around the target token count for each version. We employ exponential distribution sampling for the (G) and (R) and discrete Gaussian distribution sampling for (G+R). Please refer to Appendix~\ref{app:implement} for sampling details.

\section{MergeVQ for Efficient Generation}
\label{sec:generation}
MergeVQ supports two different AR generation paradigms: (i) raster-order generation with our tailored MergeAR for KV cache compression and (ii) the random-order one that employs randomized AR generators like RandAR~\cite{cvpr2025RandAR} enhanced by our Source Recovery Model (in Sec.~\ref{sec:mergevq_rec+rep}).

\subsection{MergeAR with KV Cache Compression}
\label{sec:mergear_gen}
MergeAR exploits the intrinsic redundancy with autoregressive token sequences by dynamically pruning duplicates to accelerate raster-order generation while preserving the spatial coherence with a position-recording system.

\begin{figure}[t!]
    \vspace{-0.25em}
    \centering
    \includegraphics[width=1.0\linewidth]{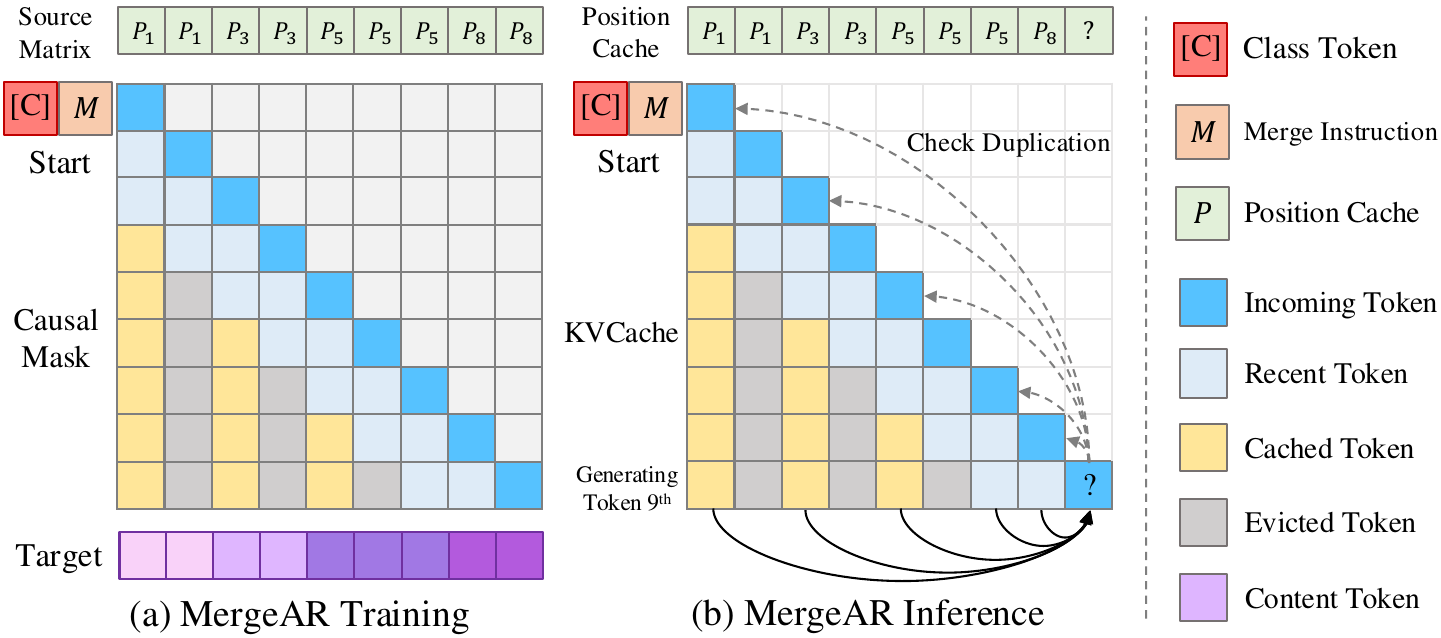}
    \vspace{-1.5em}
    \caption{\textbf{Illustration of MergeAR pipeline}. \textbf{(a)} MergeAR training with the source matrix and $K$-sparse target sequences from the MergeVQ tokenizer to build a causal mask with duplicated tokens masked out, taking a class token and a merge instruction token as the starting conditions.
    \textbf{(b)} MergeAR inference that generates $L$ tokens in the raster order with duplicated tokens detected and removed in the position and KV Caches.
    }
    \label{fig:mergeAR_pipeline}
    \vspace{-1.0em}
\end{figure}

During training, we first sample a merge ratio $r$ as in Appendix.~\ref{app:implement}, which determines the number of merged visual tokens and results in $K$ discretized tokens along with their ground-truth source matrix $S$. To regulate the level of sparsity, we introduce a Merge Instruction Token $M$, which serves as an indicator of merging extent. Using the source matrix $S$ and target $\tilde{Z}_K$, we construct a causal mask to guide the training process. Concretely, we derive a sparsity-inducing causal mask $M \in \{0, 1 \}^{L\times L}$ denoted as:
\begin{equation}
   M(i,j) =  1, \textrm{when}\  S(i,j) = 1 \ \textrm{and}  \ 1 \notin \bigcup_{k=1}^{i-1} S(k,j).
\end{equation}
This ensures each original token is represented by at most one merged token in the context. In the inference phase, we construct the KV cache similarly to the causal mask. As shown in Figure~\ref{fig:mergeAR_pipeline}, when generating the $t$-th token, MergeAR compares it against existing tokens in the KV cache. If it is a duplicate, its position will be marked as a redundant one in the Position Cache and excluded from it when the slide window moves away. Otherwise, its content token and position will be added and kept forever.



\subsection{Randomized AR with Source Recovery}
\label{sec:randar_gen}
The concurrent Randomized AR techniques (\eg, RandAR~\cite{cvpr2025RandAR}) generate tokens in arbitrary orders to improve parallelism and zero-shot generalization. Concretely, they introduce positional encoding prediction, whose objective $p_{\theta}(\mathbf{x} | \mathbf{P})$ could be formulated as:
\begin{equation}
    \prod_{n=1}^{N} p_{\theta} \left( x^{\pi(n)}_n \mid P_1^{\pi(1)}, x_1^{\pi(1)}, \dots, x_{n-1}^{\pi(n-1)}, P_n^{\pi(n)} \right),
\end{equation}
where \( x_i^{\pi(i)} \) is the \( i \)-th token in this randomly shuffled \( N \)-length sequence, and \( \pi(i) \) denotes its original position in raster order. We then insert a positional instruction token \( P_i^{\pi(i)} \) before each image token \( x_i^{\pi(i)} \).
MergeVQ can also smoothly employ randomized generation with its Source Recovery Model (Sec.~\ref{sec:mergevq_rec+rep}), where the $K$ quantized tokens $Z_{Kq}$ obtained in the first stage serve as target image tokens, and the predicted source $\hat{S}$ is used as the contextual information. After generating $K$ generated tokens, we invoke the source recovery model $ \mathcal{R}_{\omega} (\cdot, \cdot) $ and decoder, as in Eq.(\ref{equ:decode}) and Eq.(\ref{equ:rec}), to recover $L$-length tokens. Thus, when $S$ is inaccessible in inference, MergeVQ is able to conduct \textit{context-aware} token expansion for visual generation.

\section{Experiments}
\label{sec:exp}

\begin{table*}[t]
    \centering
    \vspace{-0.5em}
    \caption{\textbf{Comparison of self-supervised pre-training on ImageNet-1K}. The top-1 accuracy of linear probing (Lin.) and fully fine-tuning (FT) results are reported. $\ddag$ denotes using the multi-crop augmentation or additional data. We summarize the target for alignment (Align.) and reconstruction (Rec.), the pre-training epochs, the encoder architecture type, and the number of learnable parameters (\#Param) of the encoder and latent tokens (\#Tokens), where MIM and TMM denote Masked Image Modeling and Token-merge Modeling.
    }
    \vspace{-0.75em}
    \setlength{\tabcolsep}{1.3mm}
\resizebox{0.93\linewidth}{!}{
    \begin{tabular}{c|lc|ccc|ccc|cc}
    \toprule
Support       & Method                                & Date         & Align.  & Rec.     & Epochs & Encoder  & \#Param   & \#Tokens   & \multicolumn{2}{c}{Accuracy$\uparrow$} \\
Tasks         &                                       &              & Target  & Target   &        & Type     &           &            & Lin.           & FT            \\ \hline
              & BYOL~\cite{nips2020byol}              & NeurIPS'2020 & MSE     & \xmarkg  & 800    & R50-W2   & 94M       & 7$\times$7 & 75.6           & $-$           \\
Contrastive   & MoCov3~\cite{iccv2021mocov3}          & ICCV'2021    & InfoNCE & \xmarkg  & 300    & ViT-B    & 86M       & 196        & 76.7           & 83.2          \\
Pre-training  & DINO$^\ddag$~\cite{iccv2021dino}      & ICCV'2021    & CE      & \xmarkg  & 300    & ViT-B    & 86M       & 196        & 78.2           & 83.6          \\
              & DINOv2$^\ddag$~\cite{Oquab2023DINOv2} & TMLR'2024    & CE      & \xmarkg  & 1000   & ViT-B    & 86M       & 196        & \bf{84.5}      & \bf{85.7}     \\ \hline
              & BEiT~\cite{iclr2022BEiT}              & ICLR'2022    & \xmarkg & DALLE    & 800    & ViT-B    & 86M       & 196        & 56.7           & 83.2          \\
              & iBOT$^\ddag$~\cite{zhou2021ibot}      & ICLR'2022    & CE      & EMA      & 800    & ViT-B    & 86M       & 196        & 76.0           & 84.0          \\
              & MAE~\cite{cvpr2022MAE}                & CVPR'2022    & \xmarkg & RGB      & 1600   & ViT-B    & 86M       & 196        & 68.0           & 83.6          \\
MIM           & SimMIM~\cite{cvpr2022simmim}          & CVPR'2022    & \xmarkg & RGB      & 800    & ViT-B    & 86M       & 196        & 67.9           & 83.8          \\
Pre-training  & CAE~\cite{ijcv2023cae}                & IJCV'2023    & \xmarkg & DALLE    & 1600   & ViT-B    & 86M       & 196        & 70.4           & 83.6          \\
              & PeCo~\cite{dong2021peco}              & AAAI'2023    & \xmarkg & VQVAE    & 800    & ViT-B    & 86M       & 196        & $-$            & \bf{84.5}     \\
              & A$^2$MIM~\cite{icml2023a2mim}         & ICML'2023    & \xmarkg & RGB      & 800    & ViT-B    & 86M       & 196        & 68.8           & 84.2          \\
              & I-JEPA~\cite{cvpr2023ijepa}           & CVPR'2023    & \xmarkg & RGB      & 600    & ViT-B    & 86M       & 196        & \bf{72.9}      & $-$           \\
              & EVA-02~\cite{cvpr2024EVA02}           & CVPR'2024    & \xmarkg & EVA-CLIP & 300    & ViT-B    & 86M       & 196        & $-$            & 84.0          \\ \hline
              & ViT-VQGAN~\cite{ICLR2021VIT-VQGAN}    & ICLR'2022    & \xmarkg & RGB      & 100    & VIM-Base & 650M      & 1024       & 65.1           & $-$           \\
              & MaskGIT~\cite{CVPR2022maskgit}        & CVPR'2022    & \xmarkg & RGB      & 200    & BERT     & 227M      & 256        & 57.4           & $-$           \\
Generative    & LlamaGen~\cite{NIPS2024LLaMAGen}      & NeurIPS'2024 & \xmarkg & RGB      & 40     & CNN      & 72M       & 1024       & 47.6           & $-$           \\
              & Titok-B~\cite{nips2024titok}          & NeurIPS'2024 & \xmarkg & VQGAN    & 200    & Titok-B  & 86M       & 64         & 53.9           & $-$           \\
              & REPA~\cite{Yu2024REPA}                & ICLR'2025    & DINOv2  & Velocity & 100    & SiT-L/2  & 458M      & 1024       & \bf{71.1}      & $-$           \\ \hline
              & MAGE-C~\cite{cvpr2023MAGE}            & CVPR'2023    & InfoNCE & VQGAN    & 1600   & ViT-B    & 24+86M    & 196        & 78.2           & 82.9          \\
Generative \& & DiGIT~\cite{nips2024digit2}           & NeurIPS'2024 & DINOv2  & RGB      & 200    & ViT      & 219M      & 256        & 71.7           & $-$           \\
\rowcolor[HTML]{CFEFFF}\cellcolor{white}Pre-training & \bf{MergeVQ (G+R)} & \bf{Ours} & DINOv2 & RGB+TMM & 270 & Hybrid & 63M & 144    & 77.9           & 82.0          \\
\rowcolor[HTML]{CFEFFF}\cellcolor{white}             & \bf{MergeVQ (R)}   & \bf{Ours} & DINOv2 & RGB+TMM & 300 & ViT-B  & 86M & 36     & \bf{79.8}      & \bf{84.2}     \\
    \bottomrule
    \end{tabular}
    }
    \label{tab:in1k_ssl}
    \vspace{-0.75em}
\end{table*}

\subsection{Implementation Details}
\textbf{Visual Tokenizer Setup.}
We offer three MergeVQ versions for visual representation learning and generation: MergeVQ (G) for pure generation, MergeVQ (G+R) for both generation and representation, and MergeVQ (R) for representation learning only. As detailed in Appendix~\ref{app:network}, we present three architectures of these versions with the latent embedding dimension of 512, whose encoders have 63M, 62M, and 86M parameters. As discussed in Sec.~\ref{sec:mergevq_rec+rep}, we apply the hybrid model that contains 4 and 5 hierarchical stages of ResNet blocks~\cite{cvpr2016resnet} with 12-layer of ToMe Attention blocks~\cite{iclr2022ToMe} at the last stage for the encoder networks in MergeVQ (G) and MergeVQ (R+G), as well as LFQ layer \cite{ICLR2024magvit2} with the dimension of 18. The corresponding decoder shares a similar architecture as encoders without ToMe modules. For fair comparisons, MergeVQ (R) adopts ViT-B~\cite{iclr2021ViT} with random initialization as encoder but still adopts an identical decoder and LFQ as MergeVQ (G+R).
As for the token number after quantization, the raw output numbers of the three versions are 1024, 256, and 256, and we merge them to 256, 144, and 36 tokens during training and inference.
All versions are trained by AdamW optimizer~\cite{iclr2019AdamW} with $(\beta_1, \beta_2)$ of $(0.5, 0.9)$, a default learning rate of $1e-4$, and a total batch size of 256 for 270$\sim$300 epochs on ImageNet-1K without annotations.
As for reconstruction, models are trained in $256\times 256$ resolutions with a combination of $\ell_i$ reconstruction loss, GAN loss, perceptual loss, entropy penalty, commitment loss, and LeCAM regularization as MAGVITv2, combined with our proposed source recovery loss $\mathcal{L}_{\textrm{src}}$ and alignment loss $\mathcal{L}_{[CLS]}$.

\textbf{Visual Generator Setup.}
Following LlamaGen~\cite{NIPS2024LLaMAGen} and the concurrent work RandAR~\cite{cvpr2025RandAR}\footnote{More studies of MergeAR and the combination of MergeVQ with concurrent AR works~\cite{Yu2024RAR, cvpr2025RandAR} will be updated in the arXiv preprint.}, we conduct three versions of AR generators with MergeVQ tokenizers: MergeVQ with vanilla LlamaGen for classical raster-order generation, MergeVQ with MergeAR (built upon LlamaGen) for efficient generation, and MergeVQ with RandAR for random-order generation. As for the third version, it requires the pre-trained Source Recovery module to predict the source matrix with the generated sequences as mentioned in Sec.~\ref{sec:randar_gen}, which can be a 2-layer standard Transformer decoder with 512 embedding dimensions at 7M parameters.
We adopt LlamaGen-L as the generator architecture, which is a 24-layer Transformer decoder~\cite{Radford2019GPT2} in LLaMA-based architecture~\cite{Touvron2023LLaMA} and trained by AdamW optimizer~\cite{iclr2019AdamW} with a weight decay of 0.05, a basic learning rate of $4\times 10^{4}$, and a batch size of 1024 for 300 epochs.
View Appendix~\ref{app:stage_2} for more details.

\subsection{Self-supervised Pre-training}
\label{sec:exp_ssl}

We evaluated self-supervised pre-trained models by linear probing (Lin.)~\cite{cvpr2022MAE} and end-to-end fine-tuning (FT)~\cite{iclr2022BEiT} protocols on ImageNet-1K.
Table~\ref{tab:in1k_ssl} shows that MergeVQ variants substantially outperform prior models like BYOL, MoCov3, and DINOv2 in performance and efficiency, notably with fewer tokens achieving superior accuracy. MergeVQ (R), which focuses on representation learning, achieves impressive results with only 36 tokens. With fewer tokens than DINOv2 (196), MergeVQ (R) achieves 79.8\% Lin. accuracy and 84.2\% FT accuracy, leveraging a flexible and discriminative latent space for both efficiency and performance. 
MergeVQ (G+R) performs slightly lower than MergeVQ (R) due to its inclusion of generation alongside representation learning, highlighting the trade-off between tasks, which require more tokens, and pretraining, which benefits from coarse-grained latent. Despite this, MergeVQ (G+R) remains competitive, reaching 77.9\% of Lin. and 82.3\% of FT, demonstrating competitive results while handling both generative and representation objectives.

\begin{table}[htb]
    \centering
    \caption{\textbf{Comparison of reconstruction on 256$\times$256 ImageNet-1K} with reconstruction FID (rFID) of VQ tokenizers. We sum up the types, sizes, and dims of the codebook with its usage ratio. Ratio and \#Tokens denote the downsampling rate and token number.
    }
    \vspace{-0.75em}
    \setlength{\tabcolsep}{0.5mm}
\resizebox{1.0\linewidth}{!}{
    \begin{tabular}{l|cccc|ccc}
    \toprule
Method                                       & \multicolumn{4}{c|}{VQ Codebook}           & Ratio & \#Tokens     & rFID         \\
                                             & Type    & Size     & Dim & Usage$\uparrow$ &       & $\downarrow$ & $\downarrow$ \\ \hline
Taming-VQGAN~\cite{cvpr2021vqgan}            & Cluster & $2^{10}$ & 256 & 49\%            & 16    & $16^{2}$     & 7.94         \\
SD-VQGAN~\cite{cvpr2022ldm}                  & Cluster & $2^{10}$ & 4   & $-$             & 16    & $16^{2}$     & 5.15         \\
RQ-VAE~\cite{CVPR2022RQVAE}                  & Cluster & $2^{14}$ & 256 & $-$             & 16    & $16^{2}$     & 3.20         \\
MaskGIT~\cite{CVPR2022maskgit}               & Cluster & $2^{10}$ & 256 & $-$             & 16    & $16^{2}$     & 2.28         \\
LlamaGen~\cite{NIPS2024LLaMAGen}             & Cluster & $2^{14}$ & 8   & 97\%            & 16    & $16^{2}$     & 2.19         \\
TiTok-L-32~\cite{nips2024titok}              & Cluster & $2^{12}$ & 16  & $-$             & $-$   & 32           & 2.21         \\
TiTok-B-64~\cite{nips2024titok}              & Cluster & $2^{12}$ & 12  & $-$             & $-$   & 64           & 1.70         \\
VQGAN-LC~\cite{nips2024VQGANLC}              & CLIP    & $10^{5}$ & 8   & 99\%            & 16    & $16^{2}$     & 2.62         \\
VQ-KD~\cite{NIPS2024VQKD}                    & DINO    & $2^{13}$ & 32  & 100\%           & 16    & $16^{2}$     & 3.41         \\
MAGVIT-v2~\cite{ICLR2023MagViT2}             & LFQ     & $2^{18}$ & 1   & 100\%           & 16    & $16^{2}$     & 1.16         \\
OpenMAGVIT2~\cite{luo2024Open-Magvit2}       & LFQ     & $2^{18}$ & 1   & 100\%           & 16    & $16^{2}$     & 1.17         \\
MaskBiT~\cite{weber2024Maskbit}              & LFQ     & $2^{14}$ & 1   & 100\%           & 16    & $16^{2}$     & 1.37         \\
\rowcolor[HTML]{EFEFEF}\bf{MergeVQ (R)}      & LFQ     & $2^{18}$ & 1   & 86\%            & 16    & 144          & 4.67         \\
\rowcolor[HTML]{CFEFFF}\bf{MergeVQ (G+R)}    & LFQ     & $2^{18}$ & 1   & 99\%            & 16    & 144          & 1.48         \\
\rowcolor[HTML]{CFEFFF}\bf{MergeVQ (G+R)}    & LFQ     & $2^{18}$ & 1   & 99\%            & 16    & 256          & \bf{1.12}    \\ \hline
ViT-VQGAN~\cite{ICLR2021VIT-VQGAN}           & Cluster & $2^{13}$ & 8   & 96\%            & 8     & $16^{2}$     & 1.28         \\
OmiTokenizer~\cite{wang2024Omnitokenizer}~   & Cluster & $2^{13}$ & 8   & $-$             & 8     & $16^{2}$     & 1.11         \\
LlamaGen~\citep{NIPS2024LLaMAGen}            & Cluster & $2^{14}$ & 8   & 97\%            & 8     & $16^{2}$     & 0.59         \\
TiTok-S-128~\citep{nips2024titok}            & Cluster & $2^{12}$ & 16  & $-$             & $-$   & 128          & 1.71         \\
VQGAN-LC~\cite{nips2024VQGANLC}              & CLIP    & $10^{5}$ & 8   & 99\%            & 8     & $16^{2}$     & 1.29         \\
\rowcolor[HTML]{CFEFFF}\bf{MergeVQ (G)}      & LFQ     & $2^{18}$ & 1   & 100\%           & 8     & 256          & 1.06         \\
\rowcolor[HTML]{CFEFFF}\bf{MergeVQ (G)}      & LFQ     & $2^{18}$ & 1   & 100\%           & 8     & 1024         & \bf{0.54}    \\
    \bottomrule
    \end{tabular}
    }
    \label{tab:tokenizer_rec}
    \vspace{-1.0em}
\end{table}

\begin{table}[t]
    \centering
    \vspace{-0.25em}
    \caption{\textbf{System comparsion of class-conditional generation on 256$\times$256 ImageNet-1K}. Generation Fréchet inception distance (gFID) and inception score (IS) are reported with ADM~\cite{nips2021adm}. ``\# P" means the parameter number, step means sampling steps, and $\ddag$ denotes training tokenizers on OpenImages. Note that ``-cfg" or ``-re" denotes using classifier-free guidance or rejection sampling, and ``-384" denotes for generating images at $384\times 384$ resolutions and then resize back to $256\times 256$ for evaluation.
    }
    \vspace{-0.75em}
    \setlength{\tabcolsep}{0.3mm}
\resizebox{1.0\linewidth}{!}{
    \begin{tabular}{c|cccccc}
    \toprule
Type                                              & Tokenizer                          & Generator                                                & \# P.    & Step & gFID$\downarrow$ & IS$\uparrow$ \\ \hline
                                                  &                                    & \small{LDM-4~\cite{cvpr2022ldm}}                         & 400M     & 250  & 3.60            & 247.7        \\
                                                  &                                    & \small{UViT-L/2~\cite{cvpr2023UViT}}                     & 287M     & 250  & 3.40            & 219.9        \\
                                                  &                                    & \small{UViT-H/2~\cite{cvpr2023UViT}}                     & 501M     & 250  & 2.29            & 263.9        \\
Diff.                                             & \gray{\small{VAE$^\ddag$}}         & \small{DiT-XL/2~\cite{iccv2023DiT}}                      & 675M     & 250  & 2.27            & 278.2        \\
                                                  &                                    & \small{MDTv2-XL/2~\cite{iccv2023MDTv2}}                  & 676M     & 250  & \bf{1.58}       & \bf{314.7}   \\
                                                  &                                    & \small{SiT-XL~\cite{eccv2024SIT}}                        & 675M     & 250  & 2.06            & 270.3        \\
                                                  &                                    & \small{DiMR-XL/2R~\cite{nips2024DiMR}}                   & 505M     & 250  & 1.70            & 289.0        \\ \hline
                                                  & \gray{\small{VQGAN}}               & \small{MaskGIT~\cite{CVPR2022maskgit}}                   & 177M     & 8    & 6.18            & 182.1        \\
                                                  & \footnotesize{TiTok-B-64$^\ddag$}  & \small{MaskGIT-ViT~\cite{CVPR2022maskgit}}               & 177M     & 8    & 2.48            & 262.5        \\
Mask.                                             & \footnotesize{TiTok-S-128$^\ddag$} & \footnotesize{MaskGIT-UViT-L~\cite{cvpr2023UViT}}        & 287M     & 64   & 1.97            & 281.8        \\
                                                  & \small{MAR}                        & \small{MAR-B-cfg~\cite{NIPS2024MAR}}                     & 208M     & 100  & 2.31            & 281.7        \\
                                                  & \small{MAR}                        & \small{MAR-L-cfg~\cite{NIPS2024MAR}}                     & 479M     & 100  & \bf{1.78}       & \bf{296.0}   \\ \hline
                                                  &                                    & \small{VAR-d16~\cite{tian2024VAR}}                       & 310M     & 10   & 3.30            & 274.4        \\
VAR                                               & \small{VAR$^\ddag$}                & \small{VAR-d20~\cite{tian2024VAR}}                       & 600M     & 10   & 2.57            & 302.6        \\
                                                  &                                    & \small{VAR-d24~\cite{tian2024VAR}}                       & 1.0B     & 10   & \bf{2.09}       & \bf{312.9}   \\ \hline
                                                  & \small{VQGAN}                      & \gray{\small{GPT2~\cite{Radford2019GPT2}}}               & 1.4B     & 256  & 15.78           & 74.3         \\
                                                  & \small{VQGAN}                      & \gray{\small{GPT2-re~\cite{Radford2019GPT2}}}            & 1.4B     & 256  & 5.20            & 280.3        \\
                                                  & \small{VIT-VQGAN}                  & \small{VIM-L~\cite{ICLR2021VIT-VQGAN}}                   & 1.7B     & 1024 & 4.17            & 175.1        \\
                                                  & \small{ViT-VQGAN}                  & \small{VIM-L-re~\cite{ICLR2021VIT-VQGAN}}                & 1.7B     & 1024 & 3.04            & 227.4        \\
                                                  & \small{RQ-VAE}                     & \small{RQ-Trans.-re~\cite{CVPR2022RQVAE}}                & 3.8B     & 64   & 3.80            & 323.7        \\
                                                  & \small{MAGVIT-v2}                  & \small{MAGVIT-cfg~\cite{CVPR2023MAGVIT}}                 & 307M     & 256  & 1.78            & 319.4        \\
AR                                                & \small{LlamaGen}                   & \small{LlamaGen-L~\cite{NIPS2024LLaMAGen}}               & 343M     & 256  & 3.80            & 248.3        \\
\small{(raster)}                                  & \small{LlamaGen}                   & \footnotesize{LlamaGen-L-384~\cite{NIPS2024LLaMAGen}}    & 343M     & 576  & 3.07            & 256.1        \\
                                                  & \small{LlamaGen}                   & \small{LlamaGen-XL~\cite{NIPS2024LLaMAGen}}              & 775M     & 256  & 3.39            & 227.1        \\
                                                  & \small{LlamaGen}                   & \footnotesize{LlamaGen-XL-384~\cite{NIPS2024LLaMAGen}}   & 775M     & 576  & 2.62            & 244.1        \\
                                                  & \footnotesize{OpenMAGVIT2}         & \footnotesize{OpenMAGVIT2-B\cite{luo2024Open-Magvit2}}   & 343M     & 256  & 3.08            & 258.3        \\
                                                  & \footnotesize{OpenMAGVIT2}         & \footnotesize{Open-MAGVIT2-L\cite{luo2024Open-Magvit2}~} & 804M     & 256  & 2.51            & 271.7        \\
                                                  & \small{MaskBit}                    & \small{LlamaGen-cfg~\cite{NIPS2024LLaMAGen}}             & 305M     & 256  & \bf{1.52}       & \bf{328.6}   \\ \hline
                                                  & \gray{\small{VQGAN}}               & \small{MAGE-L}~\cite{cvpr2023MAGE}                       & 230M     & 20   & 6.93            & 195.8        \\
AR \&                                             & \gray{\small{VQGAN}}               & \small{DiGIT~\cite{nips2024digit2}}                      & 732M     & 256  & 3.39            & 206.0        \\
\rowcolor[HTML]{CFEFFF}\cellcolor{white} PT       & \small{\bf{MergeVQ (G+R)}}         & \small{LlamaGen-L~\cite{NIPS2024LLaMAGen}}               & 343M     & 256  & 3.28            & 251.6        \\
\rowcolor[HTML]{CFEFFF}\cellcolor{white}          & \small{\bf{MergeVQ (G+R)}}         & \small{\bf{MergeAR (Ours)}}                              & 343M     & 256  & 3.25            & 253.8        \\
\rowcolor[HTML]{CFEFFF}\cellcolor{white}          & \small{\bf{MergeVQ (G)}}           & \small{\bf{MergeAR (Ours)}}                              & 343M     & 1024 & \bf{3.05}       & \bf{260.9}   \\ \hline
                                                  & \gray{\small{LlamaGen}}            & \small{RandAR-L-cfg~\cite{cvpr2025RandAR}}               & 343M     & 88   & 2.55            & 288.8        \\
AR                                                & \gray{\small{LlamaGen}}            & \small{RandAR-L-cfg~\cite{cvpr2025RandAR}}               & 775M     & 88   & 2.25            & 317.8        \\
\rowcolor[HTML]{CFEFFF}\cellcolor{white} \small{(random)} & \small{\bf{MergeVQ (G+R)}} & \small{RandAR-L-cfg~\cite{cvpr2025RandAR}}               & 343M     & 64   & 2.63            & 279.5        \\
\rowcolor[HTML]{CFEFFF}\cellcolor{white}          & \small{\bf{MergeVQ (G)}}           & \small{RandAR-L-cfg~\cite{cvpr2025RandAR}}               & 343M     & 88   & \bf{2.24}       & \bf{320.4}   \\
    \bottomrule
    \end{tabular}
    }
    \label{tab:in1k_gen}
    \vspace{-1.0em}
\end{table}

\subsection{Image Generation}
\label{sec:exp_gen}

\textbf{Reconstruction.}
Table~\ref{tab:tokenizer_rec} compares the reconstruction performance of VQ-based tokenizers on $256\times 256$ ImageNet-1K. MergeVQ (G+R) achieves an effective balance between reconstruction and token efficiency (nearly a 100\%-utilized LFQ codebook with dynamic token lengths), leading to an rFID of 1.48. This outperforms methods that use larger codebooks and more tokens, such as RQ-VAE and LlamaGen. MergeVQ (G), applying the same codebook but with 256 tokens, hits an even lower rFID of 0.54, excelling in reconstruction quality. 
Overall, MergeVQ variants show high performance by optimizing codebook and token usage. While MergeVQ (G+R) slightly sacrifices rFID for handling both generation and representation, it remains competitive, highlighting the trade-off between these objectives.

\textbf{Class Conditional Generation.}
As shown in Table~\ref{tab:in1k_gen}, MergeVQ (G+R) and MergeVQ (G) stand out as competitive models. MergeVQ (G+R) uses 144 latent tokens and our MergeAR and achieves a gFID of 3.27 and an IS of 253.8 without CFG. When CFG and the concurrent RandAR generator
are applied, it improves to a gFID of 2.63 and an IS of 279.5, surpassing most AR models. On the other hand, MergeVQ (G) with MergeAR, which uses 256 tokens and 1024 steps, demonstrates even better performance, with a gFID of 3.05 and an IS of 260.9 without CFG, and achieving a gFID of 2.24 and IS of 320.4 with CFG and RandAR. By leveraging fewer tokens than several resource-intensive models (\eg, VQGAN and ViT-VQGAN with large scales), MergeVQ variants excel in class-conditional image generation by balancing generation quality and efficiency, setting a new benchmark for models in this domain. This makes MergeVQ particularly promising for real-world applications where efficiency and generation quality are both crucial. Using fewer tokens while maintaining high image quality, MergeVQ variants achieve competitive results with a more streamlined and efficient approach compared to advanced diffusion and GAN-based models.

\subsection{Ablation Study}
\label{sec:ablation}
We conduct ablation studies on ImageNet-1K.
As for tokenizers, Table~\ref{tab:ablation_tokenizer} shows that MergeVQ (G) and MergeVQ (R) could achieve the best reconstruction and pre-training performance with 256 tokens (\textit{i.e.}, adaptive downsampling instead of convolution projection) and 36 tokens (\textit{i.e.}, a small number of semantic tokens for better global alignment). MergeVQ (G+R) could well balance the reconstruction performance with the pre-training and efficiency (fewer steps and FLOPs) by 144 tokens.
As for generation, we validate these variants in Sec.~\ref{sec:generation}. As shown in Table~\ref{tab:ablation_generator}, Source Recovery is essential to restore positional information for MergeVQ (G+R) with RandAR, which could approximate the ground-truth $\mathcal{S}$ recover positions for AR generator. Table~\ref{tab:in1k_gen} and Table~\ref{tab:ablation_generator} show that KV Cache compression in MergeAR could be useful when the generated sequence is redundant, improving vanilla LlamaGen by 0.09 \vs 0.03 gFID with our MergeVQ (G) and MergeVQ (G+R).

\begin{table}[t]
    \centering
    \vspace{-0.25em}
    \caption{\textbf{Ablation of three versions of MergeVQ tokenizers} with the number of kept tokens during training for pre-training (linear probing Acc.) and reconstruction (rFID) tasks on ImageNet-1K.
    }
    \vspace{-0.75em}
    \setlength{\tabcolsep}{0.8mm}
\resizebox{1.0\linewidth}{!}{
    \begin{tabular}{c|c|cccc|c}
    \toprule
         & G                   & \multicolumn{4}{c|}{G+R}                                                                & R                 \\
\#Tokens & rFID $(\downarrow)$ & rFID $(\downarrow)$ & \# Step ($\downarrow$) & Acc. ($\uparrow$) & FLOPs $(\downarrow)$ & Acc. ($\uparrow$) \\ \hline
256      & \grow{\bf{1.41}}    & 2.15                & 64                     & 48.6              & 76.2G                & $-$               \\
196      & 1.89                & 2.53                & 49                     & 49.5              & 74.8G                & 51.2              \\
144      & 2.03                & \grow{3.07}         & \grow{36}              & \grow{51.0}       & \grow{73.4G}         & 52.5              \\
100      & 2.96                & 4.62                & 25                     & 51.2              & 72.4G                & 53.9              \\
64       & 4.74                & 6.51                & 16                     & 51.8              & 71.5G                & 54.1              \\
36       & $-$                 & 8.94                & 9                      & 52.1              & 71.7G                & \grow{\bf{54.3}}  \\
    \bottomrule
    \end{tabular}
    }
    \label{tab:ablation_tokenizer}
    \vspace{-0.5em}
\end{table}

\begin{table}[t]
    \centering
    \caption{\textbf{Ablation of main modules for MergeVQ generation} with reconstruction (rFID) and generation (gFID) evaluation.
    }
    \vspace{-0.75em}
    \setlength{\tabcolsep}{0.5mm}
\resizebox{1.0\linewidth}{!}{
    \begin{tabular}{c|cc|ccc}
    \toprule
Version                             & $\mathcal{R}$              & $\mathcal{G}$ & rFID & gFID & \# Token \\ \hline
(G+R)                               & Ground-truth $\mathcal{S}$ & \xmark        & 1.48 & $-$  & 144      \\
(G+R)                               &  2-layer Cross-Attention   & \xmark        & 1.71 & $-$  & 144      \\
\rowcolor[HTML]{EFEFEF}(G+R)+RandAR &  2-layer Cross-Attention   &  LlamaGen-L   & 1.71 & 2.63 & 144      \\ \hline
(G+R)+LlamaGen                      & \xmark                     &  LlamaGen-L   & $-$  & 3.28 & 256      \\
(G)+LlamaGen                        & \xmark                     &  LlamaGen-L   & $-$  & 3.14 & 1024     \\
\rowcolor[HTML]{EFEFEF}(G)+MergeAR  & \xmark                     &  LlamaGen-L   & $-$  & 3.05 & 1024     \\
    \bottomrule
    \end{tabular}
    }
    \label{tab:ablation_generator}
    \vspace{-0.5em}
\end{table}


\section{Conclusion}
\label{sec:conclusion}

This paper presents MergeVQ, a unified framework that bridges competing objectives of visual representation learning and image generation. It incorporates flexible token merging-based designs to balance compact latent space and fine-grained generation. In addition, we propose MergeAR, a KVCache compressive technique that yields considerable speed gains while retaining superior second-stage image generation ability.
Experiments show that MergeVQ achieves competitive performance in both pre-training and image generation, which highlights MergeVQ's versatility to adapt to both generative and discriminative demands.

\section*{Acknowledgement}
This work was supported in part by Chinese National Natural Science Foundation Projects U23B2054, 62276254, 62306313, the Beijing Science and Technology Plan Project Z231100005923033, Beijing Natural Science Foundation L221013, and the InnoHK program.
This work was done when Juanxi Tian interned at Westlake University. We also thank OPPO AI Center and AI Station of Westlake University for the support of GPUs.

{\small
\bibliographystyle{ieeenat_fullname}
\bibliography{main}
}

\clearpage
\renewcommand\thefigure{A\arabic{figure}}
\renewcommand\thetable{A\arabic{table}}
\setcounter{table}{0}
\setcounter{figure}{0}
\setcounter{page}{1}
\maketitlesupplementary

\newpage
\appendix


\section{Implementation Details}
\label{app:implement}

\subsection{Stage 1: MergeVQ Tokenizer}
\label{app:network}

\paragraph{Tokenizer Network.}
MergeVQ introduces hybrid encoders with self-attention blocks~\cite{iclr2021ViT} using ToMe modules~\cite{iclr2022ToMe}, built after the bottom of the pure CNN blocks (Residual modules with $3\times 3$ convolutions~\cite{cvpr2016resnet}) proposed in MAGVITv2~\cite{ICLR2024magvit2}. We provide three versions of MergeVQ tokenizers, where the G and G+R versions use the hybrid encoders, while the R version uses the vanilla ViT-B~\cite{iclr2021ViT}.
The specific network configurations, experimental settings, and training details are thoroughly described in Table~\ref{tab:mergevq_config}. The corresponding decoder shares a similar architecture as encoders except for using ToMe modules and replacing FFN with MixFFN~\cite{cvmj2022PVTv2, iclr2024MogaNet}. MergeVQ (G+R) and (G) versions initialize the parameters in the Transformer encoder with the DINOv2~\cite{Oquab2023DINOv2} pre-trained model (\textit{i.e.}, DINOv2-Base) by weight selection~\cite{iclr2023InitWeight}, while MergeVQ (R) adopts ViT-B~\cite{iclr2021ViT} without pre-training as the encoder.
Following the setup of the OpenMAGVIT2 \cite{luo2024Open-Magvit2} codebase, we also remove the gradient penalty loss and replace StyleGAN with PatchGAN as the discriminator (not employing DINO discriminator as VAR~\cite{tian2024VAR} in the current version). During training, we apply the reconstruction loss, the GAN loss, the perceptual loss, and the commitment loss, combined with the proposed source recovery loss $\mathcal{L}_{\textrm{src}}$ as Eq.~(\ref{eq:loss_src}) and the alignment loss $\mathcal{L}_{[CLS]}$ as Eq.~(\ref{eq:alignment}).

\vspace{-0.5em}
\paragraph{Source Recovery Model.}
The network details of the Source Recovery module in MergeVQ are shown in Table~\ref{tab:mergevq_generator}, where we utilize two Transformer decoder blocks to predict the source matrix $\hat{S}$ with quantized tokens $\tilde{Z}_{K}$. As for implementation, we utilize the standard Transformer decoder to compute from the $K$ quantized tokens (as KV embeddings) and $L$ learnable recovery queries (as query position embeddings) similar to Maskformer~\cite{nips2021Maskformer}.
As for MergeVQ with Randomized AR generators, we further fine-tuned this module with the learned generator after stage-2 training. Although the Source Recovery model was optimized in the stage-1 training (regarded as the contextual representation learning task), the additional fine-tuning could further enhance its robustness and generalization abilities for the generation task.
As for MergeAR, it does not require the assistance of the Source Recovery module, which achieves speed-up by the proposed KV Cache compression.

\begin{table}[ht]
    \setlength{\tabcolsep}{0.3mm}
    \centering
    \caption{
    Configuration of the network, weights of loss functions, and training settings for the three versions of MergeVQ tokenizers on ImageNet-1K. Note that the network designs are specified for the encoder, and the reported FLOPs are calculated for the encoder and decoder with ToMe~\cite{iclr2022ToMe} on $256\times 256$ resolutions.
    }
    \vspace{-0.5em}
\resizebox{1.0\linewidth}{!}{
\begin{tabular}{l|ccc}
    \toprule
Settings                    & G                  & G+R                      & R           \\ \hline
Base channels               & 64                 & 64                       & 768         \\
CNN Stage number            & 4                  & 5                        & $-$         \\
Channel multiplier          & [1, 2, 4, 8]       & [1, 1, 2, 4, 8]          & [1]         \\
Residual Blocks             & [4, 4, 4, 4]       & [4, 4, 4, 4, 4]          & $-$         \\
Attention Blocks            & [0, 0, 0, 12]      & [0, 0, 0, 0, 12]         & [12]        \\
Downsampling ratio          & [1, 1/2, 1/4, 1/8] & [1, 1/2, 1/4, 1/8, 1/16] & [1/16]      \\
Vocabulary size             &                    & $2^{18}$                 &             \\
Keep token number           & 256                & 144                      & 36          \\ \hline
Discriminator loss          &                    & 0.8                      &             \\
Perceptual loss             &                    & 0.7                      &             \\
LeCam regularization        &                    & 0.01                     &             \\
L2 reconstruction           &                    & 1.0                      &             \\
Commitment loss             &                    & 0.25                     &             \\
LFQ Entropy loss            &                    & 0.1                      &             \\
Source recovery loss        & 0.5                & 0.5                      & 1.0         \\
Alignement loss             & 0.1                & 1.0                      & 1.0         \\ \hline
Optimizer                   &                    & AdamW                    &             \\
($\beta_1$, $\beta_2$)      &                    & (0.5, 0.9)               &             \\
Weight decay                &                    & 0.0                      &             \\
Training epochs             & 270                & 270                      & 300         \\
Base learning rate          &                    & 1e-4                     &             \\
Batch size                  &                    & 256                      &             \\
LR scheduler                & Step               & Step                     & Cosine      \\
Gradient clipping           & $-$                & $-$                      & 5.0         \\
EMA decay                   &                    & 0.999                    &             \\ \hline
\#Param. of Encoder         & 62.3M              & 62.7M                    & 86.6M       \\
FLOPs of Encoder            & 97.5G              & 46.4G                    & 9.5G        \\
\#Param. of Decoder         & 82.8M              & 83.4M                    & 83.4M       \\
FLOPs of Decoder            & 169.2G             & 65.6G                    & 65.6G       \\
    \bottomrule
    \end{tabular}
    }
    \label{tab:mergevq_config}
    \vspace{-0.5em}
\end{table}

\vspace{-0.5em}
\paragraph{Token Merge Module.}
Following the design principle of ToMe \cite{iclr2022ToMe}, The Token Merge Module reduces the number of tokens to improve efficiency while maintaining accuracy. Unlike token pruning, which drops tokens, ToMe combines similar tokens into one representation, preserving more information and reducing accuracy loss, making it a practical, lightweight approach for both inference and training.
Specifically, the token merging process consists of the following four steps:  
\begin{itemize}
    \item Tokens are evenly divided into two groups, \( A \) and \( B \), based on their odd or even positions.  
    \item Each token in \( A \) is paired with most similar token in \( B \).  
    \item The \( r \) most similar pairs are selected for merging.  
    \item The features of tokens in these pairs are averaged to create a single representation.
\end{itemize}
Token similarity is determined using the keys (\( K \)) from the self-attention mechanism, with metrics like cosine similarity or dot product to measure similarity between tokens in \( A \) and \( B \). Since merged tokens represent multiple originals, attention computation is affected. To address this, the softmax attention scores are adjusted by adding \( \log s \), where \( s \) is the token size, ensuring merged tokens have the correct influence and maintain consistency in representation.

\begin{equation}
A = \text{softmax}\left(\frac{QK^\top}{\sqrt{d}} + \log s\right),
\end{equation}
where $A$ denotes the attention weight matrix, $Q$ denotes the query matrix, derived from the input tokens, $K$ denotes the key matrix, also derived from the input tokens, $\log s$ denotes the size adjustment term, where 
$s$ represents the length of the sequence, indicating the number of original patches it represents after merging.
In practice, two types of merging schedules are provided: (1) \textbf{Linearly Decreasing Schedule}. The number of merged tokens linearly decreases as the layer depth increases. (2) \textbf{Square Decreasing Schedule}. The number of merged tokens decreases as the layer depth increases in the squared schedule. 
These strategies allow flexibility in balancing computational efficiency and model performance. We choose the square decreasing schedule.

\subsection{Stage 2: MergeVQ Generation}
\label{app:stage_2}
We conducted raster-order and random-order autoregressive (AR) generation experiments based on LlamaGen~\cite{NIPS2024LLaMAGen} (modified by OpenMAGVIT2~\cite{luo2024Open-Magvit2}) and RandAR~\cite{cvpr2025RandAR}.
Using the LlaMA-based architecture, we adopted 2D RoPE, SwiGLU, and RMSNorm, which have been shown to be effective in previous works and thoroughly described in Table~\ref{tab:mergevq_generator}. The class embedding, indexed from a set of learnable embeddings, serves as the starting token. As for MergeAR, we also insert a Merge Instruction token, which is a learnable embedding token with a given merge number. For MergeVQ with RandAR~\cite{cvpr2025RandAR}, the classifier-free guidance (CFG)~\cite{Ho2022CFG} with a linear sampling schedule is adopted as randomized AR variants~\cite{Yu2024RAR, cvpr2025PAR}, where the optimal CFG weight is determined through a sweep with a step size of 0.1 across all
methods.

\begin{table}[t]
    \setlength{\tabcolsep}{1.2mm}
    \centering
    \caption{
    Configuration of generators and Source Recovery model in MergeVQ or MergeAR for image generation on ImageNet-1K.
    }
    \vspace{-0.5em}
\resizebox{1.0\linewidth}{!}{
\begin{tabular}{l|ccc}
    \toprule
Settings               & LlamaGen-L        & RandAR-L          & Source Recovery   \\ \hline
Base channels          & 1024              & 1024              & 384               \\
Depth                  & 24                & 24                & 2                 \\
Attention heads        & 16                & 16                & 8                 \\
FFN dimension          & 4096              & 4096              & 1536              \\
Dropout                & 0.1               & 0.1               & 0                 \\
Mask schedule          & Arccos            & Arccos            & $-$               \\
Label smoothing        & 0.1               & 0.1               & $-$               \\
\# Parameter           & 343M              & 343M              & 7M                \\ \hline
Optimizer              & AdamW             & AdamW             & AdamW             \\
($\beta_1$, $\beta_2$) & (0.9, 0.99)       & (0.9, 0.95)       & (0.9, 0.95)       \\
Weight decay           & 5e-2              & 5e-2              & 1e-2              \\
Training epochs        & 300               & 300               & 5 (optional)      \\
Base learning rate     & $4\times 10^{-4}$ & $4\times 10^{-4}$ & $1\times 10^{-4}$ \\
Batch size             & 1024              & 1024              & 256               \\
LR scheduler           & Step              & Step              & Step              \\
Gradient clipping      & 1.0               & 1.0               & $-$               \\
    \bottomrule
    \end{tabular}
    }
    \label{tab:mergevq_generator}
    \vspace{-0.5em}
\end{table}

\subsection{Merge Ratio Sampling Strategy}
\label{app:merge_sampling}
Although our proposed MergeVQ framework can target certain tasks (representation learning or generation) by choosing a certain merge ratio, it can also benefit from a wide range of merge ratios, a kind of data augmentation that enhances the generation abilities with dynamic merge ratios.
During training, we determine the corresponding merge ratio $r$ by sampling the number of tokens retained, focusing on a range around the target token count for each version. For the versions with 256 and 36 semantic tokens, we use a discrete exponential distribution to sample the varying token counts as follows:
\begin{equation}
    P(T = k) = (1 - \exp({-\lambda})) \exp({-\lambda k}),
\end{equation}
where $T$ represents the variation in the number of tokens with the index $k\ge 0$.
As for the G and R versions, the number of retained tokens is $K = (16-T)^2$ and $(6+T)^2$.
As for the (R+G)-version in Figure~\ref{fig:merge_ratio_sampling}, we use a discrete Gaussian distribution for sampling. 
\begin{equation}
    P(T = k) = \frac{\exp({-\frac{(k - \mu)^2}{2\sigma^2}})}{Z}, \quad k \in \mathbb{Z},
\end{equation}
where retained semantic tokens in the training are $(12+T)^2$.

\subsection{Evaluation of Representation Learning}
\label{app:evaluation}
As for the linear probing protocol, we follow MAE variants \citep{cvpr2022MAE, ijcv2023cae} to evaluate the linear classification performance in the latent token space of trained models. Specifically, we train a parameter-free BN layer and a linear layer for 90 epochs using AdamW optimizer with a batch size of 1024, the Cosine annealing learning rate scheduler, where the initial learning rate is set to $1\times 10^{-3}$.
As for the fine-tuning protocol, we follow SimMIM variants \citep{cvpr2022simmim, icml2023a2mim} to fully fine-tune the pre-trained encoder for 100 epochs with AdamW optimizer and a batch size of 1024, which requires advanced augmentations and training strategies for modern architectures~\cite{iccv2019cutmix, eccv2022AutoMix}.
The MergeVQ tokenizers use all tokens (\textit{i.e.}, not applying ToMe) for both the linear probing and full fine-tuning evaluations in Table~\ref{tab:in1k_ssl}, which could yield better performance with all vision tokens in the encoder. Meanwhile, the MergeVQ (R) tokenizer utilizes 144 tokens for reconstruction evaluation in Table~\ref{tab:tokenizer_rec}. We found that it will degenerate rFID and cause more computational overhead when using all tokens because of the distribution gaps between 36-token pre-training and 256-token evaluation.

\section{More Experiment Results}
\label{app:result}
\begin{figure*}[t]
    \vspace{-0.5em}
    \centering
    \includegraphics[width=0.98\textwidth]{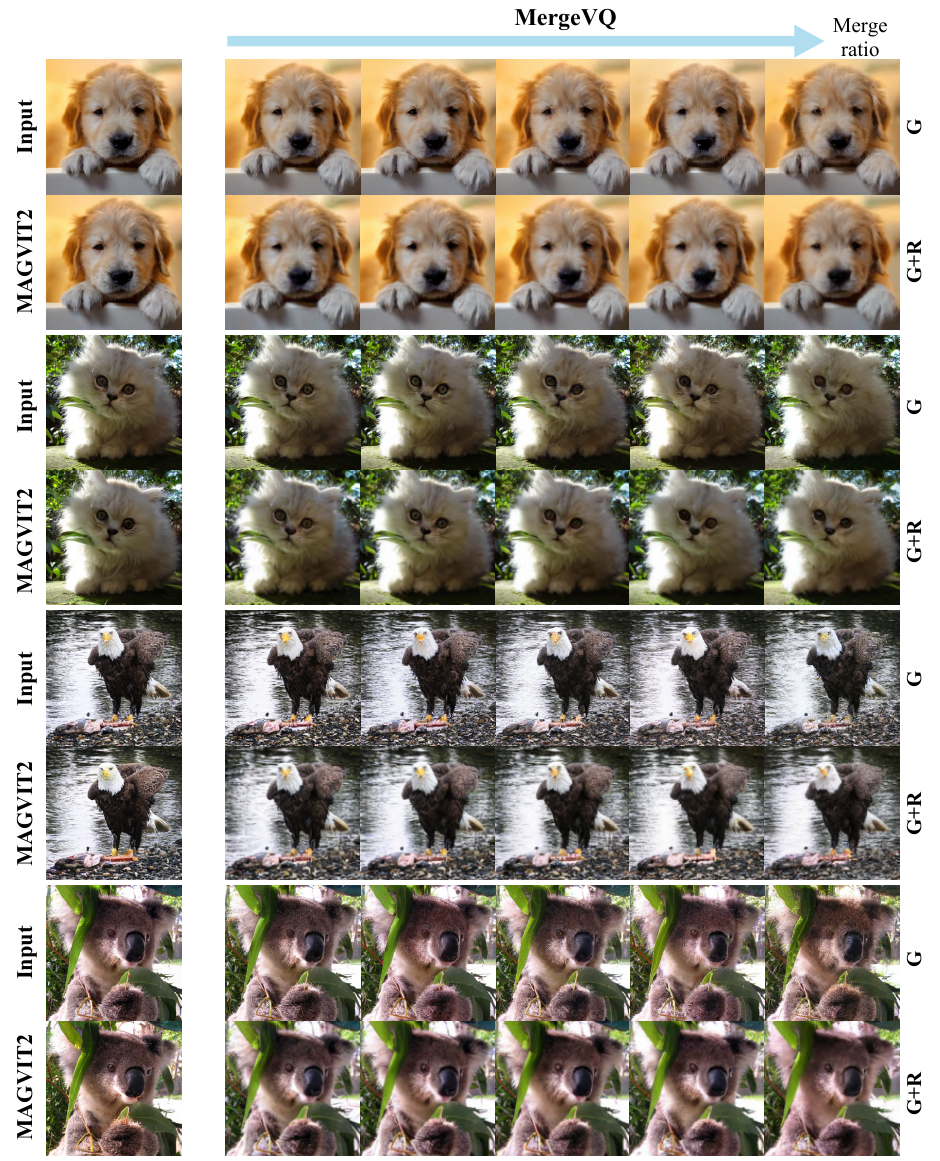}
    \vspace{-0.5em}
    \caption{\textbf{Visualization of tokenizer reconstruction on ImageNet-1K.}
    We conducted reconstruction experiments with our G version using 1024, 576, 400, 256, and 144 tokens and with our G+R version using 256, 196, 144, 100, 64, and 36 tokens. The reconstruction results are shown in the figure. As the number of retained tokens increases, the reconstruction becomes more realistic.
    }
    \label{fig:app_vis_rec}
    \vspace{-1.0em}
\end{figure*}

\begin{figure*}[t]
    \vspace{-0.5em}
    \centering
    \includegraphics[width=0.995\textwidth]{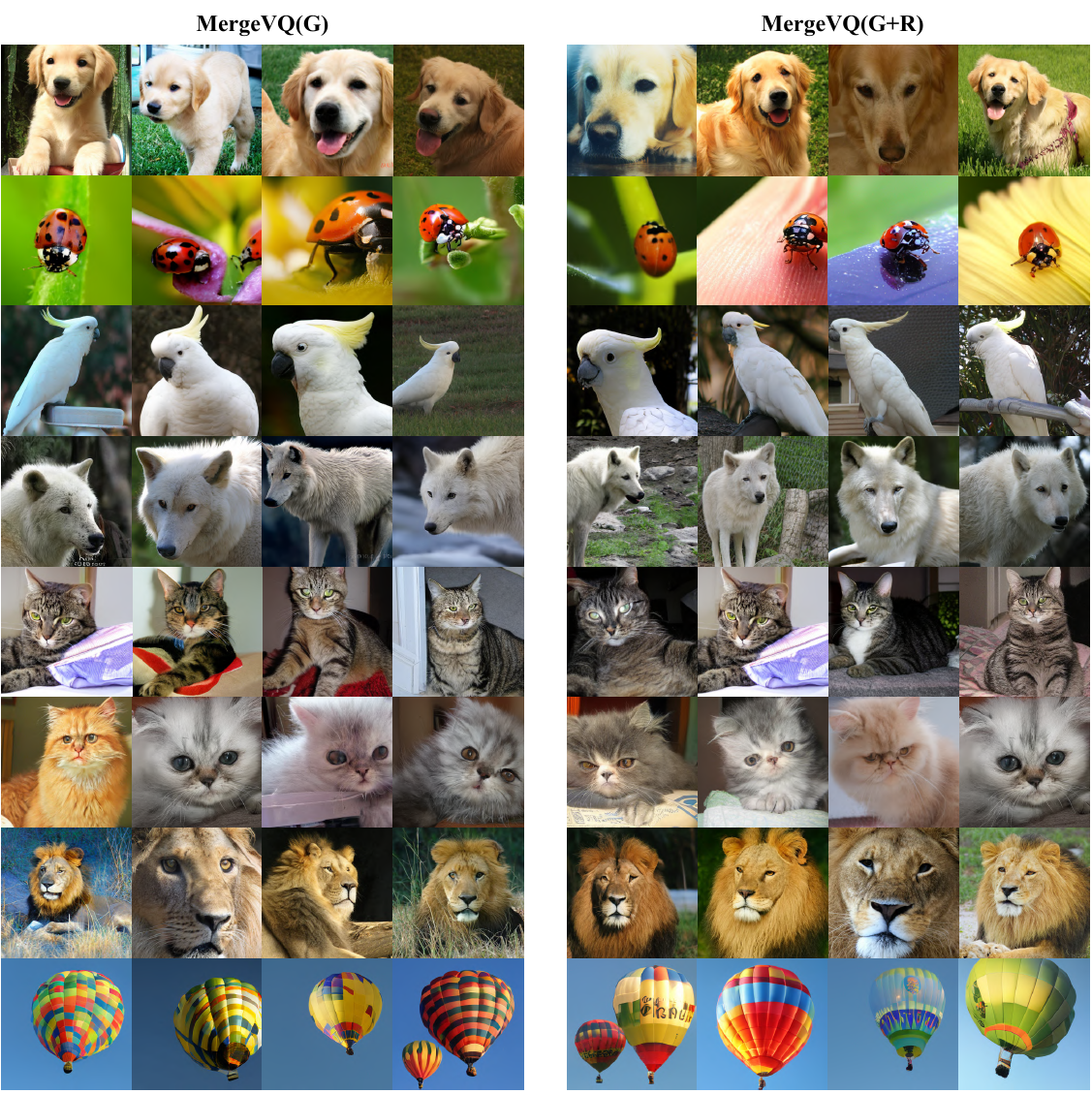}
    \vspace{-0.5em}
    \caption{\textbf{Visualization of class conditional generation} with MergeVQ variants on ImageNet-1K. The G version performs generation on 256 tokens, and the G+R version performs generation on 144 tokens.
    }
    \label{fig:app_vis_gen}
    \vspace{-1.0em}
\end{figure*}

We evaluate the reconstruction of MergeVQ (G) and MergeVQ (G+R) tokenizers at different merging ratios~\ref{fig:app_vis_rec}. The specific results can be seen in the figure, where we compare our experimental results with those of MAGVIT2~\cite{ICLR2024magvit2}.
We also visualize the generation results of MergeVQ variants in Figure~\ref{fig:app_vis_gen}, where the reconstruction quality progressively improves as the merge ratio decreases. The G+R version also achieves competitive results with 144 tokens.

\end{document}